\documentclass[
twocolumn,
]{ceurart}

\usepackage{listings}
\usepackage{afterpage}

\usepackage{tikz}
\usepackage{pgfplots}
\pgfplotsset{width=\linewidth}

\usepackage{tcolorbox}
\usepackage{xcolor}
\usepackage{lipsum}
\usepackage{makecell}
\usepackage{enumitem}
\setlist[itemize]{itemsep=0pt,parsep=0pt}

\newtcolorbox{styledtext}[2][]{%
  colback=white,
  colframe=black,
  fonttitle=\bfseries,
  title=#2,
  label=#1,
}
\newtcolorbox{styledtexttt}[2][]{%
  colback=white,
  colframe=black,
  fonttitle=\bfseries,
  fontupper=\sffamily,
  title=#2,
  label=#1,
}

\newcommand{\sans}[1]{{\fontfamily{SourceSansPro-TLF}\selectfont #1}}

\usepackage{colortbl}
\usepackage{array}

\definecolor{green0}{RGB}{255,255,255}   
\definecolor{green1}{RGB}{240,248,240}   
\definecolor{green2}{RGB}{220,240,220}   
\definecolor{green3}{RGB}{180,220,180}   
\definecolor{green4}{RGB}{140,200,140}   
\definecolor{green5}{RGB}{100,180,100}   
\definecolor{green6}{RGB}{60,160,60}     

\newcommand{\heatcell}[1]{%
  \ifnum#1<10\relax\cellcolor{green0}#1\fi%
  \ifnum#1>9\ifnum#1<20\relax\cellcolor{green1}#1\fi\fi%
  \ifnum#1>19\ifnum#1<30\relax\cellcolor{green2}#1\fi\fi%
  \ifnum#1>29\ifnum#1<50\relax\cellcolor{green3}#1\fi\fi%
  \ifnum#1>49\ifnum#1<70\relax\cellcolor{green4}#1\fi\fi%
  \ifnum#1>69\ifnum#1<85\relax\cellcolor{green5}#1\fi\fi%
  \ifnum#1>84\relax\cellcolor{green6}#1\fi%
}
\newcommand{\heatcellcolor}[1]{%
  \ifnum#1<10\relax\cellcolor{green0}\fi%
  \ifnum#1>9\ifnum#1<20\relax\cellcolor{green1}\fi\fi%
  \ifnum#1>19\ifnum#1<30\relax\cellcolor{green2}\fi\fi%
  \ifnum#1>29\ifnum#1<50\relax\cellcolor{green3}\fi\fi%
  \ifnum#1>49\ifnum#1<70\relax\cellcolor{green4}\fi\fi%
  \ifnum#1>69\ifnum#1<85\relax\cellcolor{green5}\fi\fi%
  \ifnum#1>84\relax\cellcolor{green6}\fi%
}

\newcommand{\pro}{\textsc{ProverbIT}\xspace}
\usepackage{xspace}

\usepackage{breakurl}

\definecolor{tableblue}{rgb}{0.678, 0.847, 0.902} %
\definecolor{tableorange}{rgb}{1.0, 0.7, 0.5} %
\DeclareRobustCommand{\legendsquare}[1]{%
  \textcolor{#1}{\rule{3ex}{1.5ex}}%
}

\begin{document}

\copyrightyear{2025}
\copyrightclause{Copyright for this paper by its authors.
  Use permitted under Creative Commons License Attribution 4.0
  International (CC BY 4.0).}

\conference{CLiC-it 2025: Eleventh Italian Conference on Computational Linguistics, September 24 — 26, 2025, Cagliari, Italy}

\title{Easy to Complete, Hard to Choose: Investigating LLM Performance on the ProverbIT Benchmark}
\date{May 2025}

\author[1]{Enrico Mensa}[%
email=enrico.mensa@unito.it ,
]
\cormark[1]
\fnmark[1]
\author[2]{Lorenzo Zane}[%
email=lorenzozane98@gmail.com,
]
\fnmark[1]
\author[1]{Calogero J. Scozzaro}[%
email=calogerojerik.scozzaro@unito.it,
]
\author[1]{Matteo {Delsanto}}[%
email=matteo.delsanto@unito.it,
]
\author[2]{Tommaso Milani}[%
email=milani.tommaso2004@gmail.com,
]
\author[1]{Daniele P. Radicioni}[%
email=daniele.radicioni@unito.it,
]

\address[1]{Department of Computer Science, University of Turin, Turin, Italy}
\address[2]{Independent Researcher}




\cortext[1]{Corresponding author.}
\fntext[1]{These authors contributed equally.}

\begin{abstract}
    Large Language Models (LLMs) have transformed computational linguistics and achieved remarkable performance across numerous natural language processing tasks, yet significant gaps persist in understanding how these systems process culturally embedded linguistic expressions. This paper introduces \pro, a novel Italian benchmark comprising $100$ multiple-choice questions designed to evaluate LLMs' ability to complete Italian proverbs. We assess $13$ frontier models, including Large Reasoning Models (LRMs) and traditional LLMs, across three tasks: proverb completion, multiple-choice selection with correct answers, and multiple-choice selection without correct answers. Our evaluation reveals surprising results: while nearly all models demonstrate knowledge of the proverbs through successful completion tasks, performance drops dramatically when transitioning to multiple-choice formats without correct answers, with even state-of-the-art reasoning models showing substantial degradation. Through detailed Chain-of-Thought analysis of two LRMs, we uncover that models exhibit a strong bias toward selecting literal synonyms and frequently mention correct proverb endings during reasoning without successfully identifying their absence from the given options. These findings suggest that current LLMs rely heavily on memorized patterns rather than deeper semantic understanding of culturally grounded expressions, highlighting important limitations in their reasoning capabilities for figurative language comprehension.
\end{abstract}


\maketitle

\section{Introduction}

The emergence of Large Language Models (LLMs) has revolutionized the natural language processing landscape across diverse domains, from machine translation and text summarization to code generation and complex reasoning tasks~\cite{lewkowycz2022solving}. While these models demonstrate remarkable capabilities in handling sophisticated linguistic phenomena~\cite{chang2024survey}, significant gaps persist in our comprehension of how these systems process culturally embedded linguistic expressions~\cite{fornaciari2024hard}.

Proverbs present an interesting testbed for language model evaluation. 
Informally stated, a proverb is a short, commonly known saying: it expresses a general truth, piece of wisdom, or practical advice, often based on common sense or cultural experience. The understanding of proverbs thus represents a key milestone in language proficiency, and access to the individual components of a proverb allows for the investigation of both lexical access issues and deeper semantic mechanisms.
 
These well-established expressions should be trivial for models trained on vast text corpora, as they represent highly frequent patterns that are ideal candidates for next-token prediction. A model encountering `\sans{Better late...}' should effortlessly complete it with `\sans{than never}' through simple pattern recognition. However, the challenge becomes arguably more complex when models are presented with multiple plausible proverb endings in a multiple-choice format. This shifts the task from automatic completion to deliberate selection, requiring the model not only to recognize the correct ending, but also to evaluate and dismiss semantically or syntactically plausible alternatives. Finally, another practically relevant question is: How are the performances impacted if we remove the correct answer among the possible choices, and provide the model with the option `\sans{None of the others}'? 
This transformation from pattern completion to discriminative reasoning may be insightful to investigate whether models are capable of grasping the underlying meaning of these cultural expressions, or solely rely on statistical co-occurrence patterns.

In this work we introduce \pro, a novel dataset comprising multiple-choice questions centered on Italian proverbs, designed to assess the reasoning capabilities of both Large Reasoning Models (LRMs)~\cite{xu2025towards} and traditional LLMs in handling culturally grounded linguistic expressions. By manually designing alternative endings for the proverbs, we can systematically examine the types of errors LLMs make and identify common failure patterns. Our investigation shows a striking paradox: while nearly all models possess knowledge of the proverbs in our dataset, performance deteriorates dramatically when moving from auto-completion to multiple-choice selection, with even state-of-the-art LRMs exhibiting substantial performance drops.

The contribution of this paper is threefold: \textit{i}) we contribute to Italian NLP benchmarks by introducing a novel dataset that addresses the under-representation of Italian in comprehensive language model evaluation resources~\cite{wu2025bitter}; \textit{ii}) we conduct a thorough evaluation across $13$ frontier models, including LLMs, LRMs, and smaller local models, providing comprehensive performance analysis on proverb completion tasks; and \textit{iii}) we investigate LRMs performance through detailed Chain-of-Thought (CoT) analysis, revealing insights into reasoning strategies and cultural language understanding mechanisms in contemporary language models.

The paper is organized as follows: Section~\ref{sec:sota} reviews the literature on LLM performance with idioms and proverbs; Section~\ref{sec:dataset} illustrates the \pro dataset constructive rationale and its features; Section~\ref{sec:evaluation} presents our evaluation of frontier LLMs on the task along with detailed error and CoT analysis; and Section~\ref{sec:conclusions} summarizes the work with a final discussion and an overview on future research directions.

\section{Related Work} 
\label{sec:sota}
Standardized benchmarks have been fundamental in evaluating the performance of LLMs across a variety of natural language processing tasks. Early efforts, such as GLUE \cite{wang2018glue} and SuperGLUE \cite{wang2019superglue}, were based on multi-task evaluation frameworks including tasks such as paraphrase detection, grammatical acceptability, and natural language inference.
More recently, benchmarking efforts have expanded into other domains, such as mathematics \cite{cobbe2021training, glazer2024frontiermath}, coding \cite{jimenez2024swe, jain2024livecodebench}, and complex logical reasoning tasks \cite{chollet2019measure, chollet2025arc}. These advances reflect the increasing demand for language models capable of handling a broader range of cognitive challenges.

Focusing on Italian, dedicated benchmark efforts have emerged to address language-specific issues and reduce reliance on translated tasks, which can introduce cultural bias and translation artifacts. Notable among these are CALAMITA \cite{attanasio2024calamita}, a collaborative and evolving benchmark initiative, and Evalita-LLM \cite{magnini2025evalita}. Both focus on tasks originally designed in Italian and include a mix of generative and multiple-choice tasks.

While these benchmarks cover a broad spectrum of tasks, figurative language phenomena such as idioms and proverbs have received comparatively little attention.
Idioms are a well-known source of complexity in natural language understanding \cite{Sag2002}, as their meanings often cannot be inferred literally and require contextual and cultural knowledge.
%
\citet{fornaciari2024hard} introduced an expert-curated English dataset for idiom detection, showing that local LLMs struggle to distinguish idiomatic from literal usage.
In the context of multilingual approaches,
\citet{tedeschi2022id10m} presented the ID10M dataset, a high-quality, automatically generated resource covering ten languages, along with a multilingual Transformer model for idiom identification.
Significant differences in LLM performance across languages and figurative types were reported by \citet{khoshtab2025comparative}, who evaluated simile and idiom interpretation under various prompt strategies. Notably, CoT prompting was particularly effective for similes in smaller models.
\citet{kim2025memorization} presented a dataset of idioms in six languages, each paired with its corresponding meaning.
The authors conducted a comprehensive evaluation of LLMs’ ability to process idioms, showing that models rely not only on memorization but also on a hybrid approach that integrates contextual cues and reasoning, suggesting that idiom understanding emerges from an interplay between internal knowledge retrieval and inference. Moreover, their results highlight a performance gap between high-resource and lower-resource languages.

Idioms have also been an area of interest in the machine translation (MT) literature, where multiple studies have explored how models translate figurative expressions across languages.
\citet{lee2023last} presented TIDE, a dataset of $512$ sentence pairs containing idioms in disambiguating contexts, with one sentence using the idiom literally and the other figuratively. They compared MT systems and language models, finding that the former consistently translate English idioms literally, while the latter are more context-aware, even though performance varies across target languages.
One strategy to improve idiom translation, particularly in smaller language models, involves the use of knowledge bases (KBs). \citet{li2024translate} introduced IDIOMKB, a multilingual idiom KB developed using LLMs, designed to enhance translation quality by providing access to idioms’ figurative meanings.
However, this approach does not preserve the cultural and stylistic nuances that make idioms so distinctive.
To address this issue, \citet{donthi2025improving} proposed two alignment-based methods that aim to identify idiomatic counterparts in the target language. Their results, based on human evaluation across multiple language pairs, show improved cross-lingual idiomatic fidelity and better preservation of cultural authenticity.

The work most closely related to ours is by \citet{liu2024multilingual}, who focused specifically on proverbs. They introduced the MAPS dataset, designed to evaluate proverb understanding within conversational contexts across six languages.  
Their evaluation of multilingual LLMs revealed that while many models “know” a limited set of proverbs, memorization does not guarantee understanding or contextual reasoning.
Models also struggled with figurative proverbs, particularly when asked to select incorrect answers instead of correct ones.
\citet{wang2025proverbs} extended the MAPS dataset to evaluate MT models and LLMs on proverb translation. Their experiments showed that LLMs generally outperform traditional MT models, confirming their superior ability to capture idiomatic nuances.

\section{ProverbIT Dataset}
\label{sec:dataset}

\subsection{Data Collection and Dataset Creation}
The \pro\footnote{The full dataset can be downloaded at \url{https://huggingface.co/datasets/emensa/proverbIT}.} dataset is composed of $100$ multi-choice questions, each regarding the completion of a specific Italian proverb. To create the dataset, we started from an initial set of $200$ common Italian proverbs~\cite{200_proverbi} from which we selected $100$ of the most commonly used. This process was carried out by three of the authors, which are all native Italian speakers. Each proverb was then manually split into its \textit{beginning} and its \textit{ending}, with the point of division determined to maintain the proverb's semantic coherence in the initial part while allowing for a clear, unambiguous completion. For each proverb, four distinct incorrect alternative endings were manually created, leveraging the following constructive rationale: 
\begin{itemize}\label{options_list}
    \item \textbf{A} is an ending that has similar sounds to the original continuation, often with an absurd/nonsensical meaning. 
    \item \textbf{B} is a non assonant literal synonym of the original ending. 
    \item \textbf{C} is the inverse of the original proverb ending, trying to maintain the assonance when possible.
    \item \textbf{D} is a tautological/trivial ending of the proverb, with no assonance.
\end{itemize}
For sake of clarity we provide an example in English for each of the aforementioned continuations. Completions for the proverb \sans{Actions speak... louder than words} could be:
\begin{itemize}[label={}]
    \item A) \sans{prouder than swords}
    \item B) \sans{at higher volume compared to speech}
    \item C) \sans{quieter than words}
    \item D) \sans{when they do}
\end{itemize}
As this example shows, the synonym ending is not built on the figurative meaning of the proverb, but it is the literal synonym of the original ending (e.g., \sans{at higher volume compared to speech} rather than \sans{beyond what words can say}).
This design was adopted to ensure that models cannot simply rely on surface-level syntactic patterns but must engage in deeper semantic and contextual reasoning to identify the absence of the correct completion. 

\subsection{Prompt}
\label{subsec:prompting}

Given each proverb in \pro, we can then fill a simple prompt template that can be submitted to the models:

\begin{styledtext}{Prompt Template (translated)}
\small
Complete the proverb exactly by choosing from the following options (which have no typing errors) indicating only the letter.\\

    [\textit{Proverb beginning}]...\\
    A) ...[\textit{Assonant ending]}\\
    B) ...[\textit{Synonym ending}]\\
    C) ...[\textit{Inverse ending}]\\
    D) ...[\textit{Trivial ending}]\\
    E) None of the other answers\\
    
    Do not add comments, the possible answers are only A, B, C, D, E.
\end{styledtext}

We specify that the proverb must be completed \textit{exactly}, and also that there are no typos in the options since we noticed that models often assume the presence of user mistakes and modify their responses based on this assumption. Since all provided endings are completely invented and thus incorrect, we expect models to always answer \sans{E) None of the other answers}. Finally we provide an Italian example [with translation] from the actual dataset.

\begin{styledtext}{Example of proverb from the dataset}%
A buon intenditor,... [To a wise man] \\
A) ...foche canore [singing seals] \\
B) ...zero chiacchiere [zero chatter] \\
C) ...molte parole [many words] \\ 
D) ...è chiaro tutto [everything is clear]\\ 
E) Nessuna delle altre risposte [None of the other answers]
\end{styledtext}
More examples can be found in the Supplementary Materials.

\section{Evaluation}
\label{sec:evaluation}


In this Section, we describe the experimental setup developed for evaluating $13$ different frontier models on the \pro benchmark, followed by an error analysis and in-depth examination of the underlying chain-of-thought processes for two LRM models.

\subsection{Experiments}
\begin{table*}[t]
  \begin{tabular}{llcc}
    \toprule
    Model  & Full Model Name & Provider & Num. Parameters \\
    \midrule
    Claude Sonnet 4  & claude-sonnet-4 & Anthropic & Undisclosed \\
    \rowcolor{tableblue!40} Claude Sonnet 4  & claude-sonnet-4-thinking & Anthropic & Undisclosed \\
    GPT 4o  & gpt-4o & OpenAI & Undisclosed \\
    \rowcolor{tableblue!40} GPT o3 & gpt-o3 & OpenAI & Undisclosed \\
    DeepSeek V3 & deepseek-chat-v3-0324 & DeepSeek & $671$B \\
    \rowcolor{tableblue!40} DeepSeek R1 & deepseek-r1-0528 & DeepSeek & $671$B \\
    Gemini 2.5 Flash & gemini-2.5-flash-preview-05-20 & Google & Undisclosed \\
    \rowcolor{tableblue!40} Gemini 2.5 Pro & gemini-2.5-pro-preview-06-05 & Google & Undisclosed \\
    \rowcolor{tableblue!40} Qwen 3 & Qwen 3-235b-a22b & QwQ & $235$B \\
    Grok 3 & grok-3-beta & xAI & Undisclosed \\
    LLama 4 Maverick & llama-4-maverick & Meta & $400$B \\
    \rowcolor{tableorange!40} Mistral Small 3.1 & mistral-small-3.1-24b-instruct & Mistral & $24$B \\ 
    \rowcolor{tableorange!40} Gemma 3 & gemma-3-27b-it & Google & $27$B \\
    \bottomrule
  \end{tabular}
    \caption{\legendsquare{tableblue!40}~Reasoning model \legendsquare{tableorange!40}~Local model \\ Detailed list of the models evaluated on the \pro benchmark.}
    \label{tab:full_names}
\end{table*}

In addition to evaluating the models on the \pro benchmark introduced in the previous Section, we also perform two ancillary tasks to assess whether the models possess knowledge of the proverbs.
We refer to the \pro benchmark as to the \textit{base} task, while the two ancillary tasks are described in the following.
\paragraph{Completion Task.} Instead of a multiple-choice approach, we ask the model to directly complete a proverb given its beginning. This task establishes if the model is familiar with the requested proverbs. The prompt used for the completion task is as follows:
\begin{styledtext}{Completion Prompt Template (translated)}
Complete the proverb exactly:\\

[\textit{Proverb beginning}]...\\

Reply with the ending only, do not add further comments.
\end{styledtext}

\paragraph{Base + true ending Task.} We add to each multiple-choice question a new option that is the true ending of the proverb. By preserving the multiple-choice format but also providing the correct ending, we expect similar results w.r.t. the completion task. 

\subsubsection{Evaluation \& Metrics}
In the \textit{base} and \textit{base + true ending} tasks we computed the \textbf{accuracy} defined as the ratio of correctly chosen options over the multiple choices. Specifically, each prompt was presented to each model three times and the final answer was determined through a majority vote between them. If no majority emerged across the three runs, the response was marked as incorrect. 

For the automatic calculation of the accuracy on the \textit{completion} task we compute the edit distance\footnote{The implementation from \url{https://docs.python.org/3/library/difflib.html} was employed.} between the provided completion and the correct ending of the proverb. As with the other tasks, each prompt is submitted three times. 
If the edit distance exceeds a threshold of $0.8$ in at least two out of three runs, we consider the answer correct.

For all tasks, a zero-shot prompting strategy was employed and all requests have been sent separately via API, specifically using the OpenRouter unified interface~\cite{openrouter2024}. For all models the temperature was left at the default OpenRouter value of 1.0 since we countered their nondeterministic nature by employing a majority vote.

\subsection{Models}



In our experiments we employed a diverse set of state-of-the-art models including both \textit{traditional} LLMs and LRMs, aiming to cover a wide range of providers. Whenever possible, we selected both a flagship LLM and its corresponding LRM from the same organization, allowing us to directly compare their performance and assess the improvements brought by the reasoning mechanism. The complete list of models and their full names can be found in Table \ref{tab:full_names}. 

From Anthropic, we evaluated Claude Sonnet 4\footnote{\url{https://www.anthropic.com/claude/sonnet}} and its reasoning variant Claude Sonnet 4 Thinking. From OpenAI, we included GPT 4o \cite{hurst2024gpt} and GPT o3.\footnote{\url{https://openai.com/index/o3-o4-mini-system-card/}} From DeepSeek, we employed DeepSeek V3 \cite{liu2024deepseek} and DeepSeek R1 \cite{guo2025deepseek}. From Google, we tested Gemini 2.5 Flash\footnote{\url{https://storage.googleapis.com/model-cards/documents/gemini-2.5-flash-preview.pdf}} and Gemini 2.5 Pro.\footnote{\url{https://storage.googleapis.com/model-cards/documents/gemini-2.5-pro-preview.pdf}}
We also included Qwen 3 \cite{qwen3technicalreport}, a model optimized for reasoning developed by QwQ, Grok 3\footnote{\url{https://x.ai/news/grok-3}} from xAI, and LLama 4 Maverick\footnote{\url{https://ai.meta.com/blog/llama-4-multimodal-intelligence/}} from Meta.
We also included two smaller models suitable for local deployment, as these are commonly used in privacy-sensitive contexts and contexts that require less computational resources. Although privacy concerns are not relevant for the \pro dataset, these models were included to ensure comprehensive evaluation coverage. In particular we tested Mistral Small 3.1\footnote{\url{https://mistral.ai/news/mistral-small-3-1}} from Mistral and Gemma 3 \cite{team2025gemma} from Google. 
Regarding models specifically trained on Italian, we preliminarily tested the Italian LLM Minerva \cite{orlando2024minerva} but found that it was unable to respond coherently, often failing to follow the requested response format (i.e., in providing a letter corresponding to a given choice). 

Given that some reasoning models require a mandatory thinking budget while others do not, we set a reasonable thinking budget of $2000$ tokens for o3, Sonnet 4, and Gemini 2.5 Pro, while DeepSeek R1 and Qwen 3 were left unlimited. 
Moreover, the first three models output only a summarization of their CoTs, while the latter two provide their complete trace. This makes DeepSeek R1 and Qwen 3 ideal candidates for the CoT analysis that we performed. We observed that only $22$ out of $600$ CoTs from these two models exceeded the $2000$-token limit, half of them resulting in an incorrect answer anyway.

%
\subsection{Results \& Discussion} 
In this Section we examine the results of the evaluation and provide a detailed discussion on the errors. 

\begin{table}[h!]
  \resizebox{1\columnwidth}{!}{
  \begin{tabular}{lc | cc}
    \toprule
    Model & Base & \makecell{Base + \\ True end.} & Complet. \\
    \midrule
    \rowcolor{tableblue!40} GPT o3 & \textbf{$86.0$} & $98.0$ & $91.0$ \\
    \rowcolor{tableblue!40} Gemini 2.5 Pro & $77.0$ & $99.0$ & $94.0$ \\
    \rowcolor{tableblue!40} DeepSeek R1 & $74.0$ & \textbf{$100.0$} & $89.0$ \\
    \rowcolor{tableblue!40} Claude Sonnet 4 & $73.0$ & $99.0$ & \textbf{$96.0$} \\
    \rowcolor{tableblue!40} Qwen 3 & $65.0$ & $94.0$ & $74.0$ \\
    GPT 4o & $64.0$ & $92.0$ & $88.0$ \\
    Claude Sonnet 4 & $46.0$ & $94.0$ & $93.0$ \\
    DeepSeek V3 & $40.0$ & $93.0$ & $92.0$ \\
    Grok 3 & $26.0$ & $95.0$ & $94.0$ \\
    Gemini 2.5 Flash & $12.0$ & $85.0$ & $67.0$ \\
    LLama 4 Maverick & $6.0$ & $75.0$ & $88.0$ \\
    \rowcolor{tableorange!40} Mistral Small 3.1 & $28.0$ & $71.0$ & $68.0$ \\ 
    \rowcolor{tableorange!40} Gemma 3 & $4.0$ & $48.0$ & $67.0$ \\
    \bottomrule
  \end{tabular}
  }
    \caption{\legendsquare{tableblue!40}~Reasoning model \legendsquare{tableorange!40}~Local model \\ Accuracy of models on the base task (the \pro benchmark) and the two ancillary tasks.}
    \label{tab:accuracy}
\end{table}

\begin{table}[!th]
\centering
  \resizebox{1\columnwidth}{!}{
\begin{tabular}{lcccc}
    \toprule
    Model & (A) & (B)& (C) & (D) \\
    \midrule
    \cellcolor{tableblue!40}GPT o3 & \heatcellcolor{5}$5.1$ & \heatcellcolor{87}$87.2$ & \heatcellcolor{7}$7.7$ & \heatcellcolor{0}$0.0$ \\
   \cellcolor{tableblue!40}Gemini 2.5 Pro & \heatcellcolor{4}$4.2$ & \heatcellcolor{87}$87.3$ & \heatcellcolor{7}$7.0$ & \heatcellcolor{1}$1.4$ \\
   \cellcolor{tableblue!40}DeepSeek R1 & \heatcellcolor{2}$2.3$ & \heatcellcolor{91}$91.9$ & \heatcellcolor{2}$2.3$ & \heatcellcolor{3}$3.5$ \\
   \cellcolor{tableblue!40}Claude Sonnet 4 & \heatcellcolor{3}$3.3$ & \heatcellcolor{85}$85.6$ & \heatcellcolor{11}$11.1$ & \heatcellcolor{0}$0.0$ \\
   \cellcolor{tableblue!40}Qwen 3 & \heatcellcolor{10}$10.6$ & \heatcellcolor{55}$55.8$ & \heatcellcolor{28}$28.3$ & \heatcellcolor{5}$5.3$ \\
   GPT 4o & \heatcellcolor{2}$2.9$ & \heatcellcolor{75}$75.0$ & \heatcellcolor{17}$17.3$ & \heatcellcolor{4}$4.8$ \\
   Claude Sonnet 4 & \heatcellcolor{1}$1.9$ & \heatcellcolor{62}$62.4$ & \heatcellcolor{30}$30.9$ & \heatcellcolor{4}$4.9$ \\
   DeepSeek V3 & \heatcellcolor{13}$13.5$ & \heatcellcolor{48}$48.1$ & \heatcellcolor{27}$27.6$ & \heatcellcolor{10}$10.8$ \\
   Grok 3 & \heatcellcolor{7}$7.3$ & \heatcellcolor{74}$74.3$ & \heatcellcolor{16}$16.5$ & \heatcellcolor{1}$1.8$ \\
   Gemini 2.5 Flash & \heatcellcolor{7}$7.3$ & \heatcellcolor{51}$51.9$ & \heatcellcolor{31}$31.3$ & \heatcellcolor{9}$9.5$ \\
   LLama 4 Maverick & \heatcellcolor{17}$17.5$ & \heatcellcolor{52}$52.3$ & \heatcellcolor{21}$21.1$ & \heatcellcolor{9}$9.1$ \\
   \cellcolor{tableorange!40}Mistral Small 3.1 & \heatcellcolor{9}$9.2$ & \heatcellcolor{41}$41.6$ & \heatcellcolor{27}$27.1$ & \heatcellcolor{22}$22.2$ \\
   \cellcolor{tableorange!40}Gemma 3 & \heatcellcolor{35}$35.5$ & \heatcellcolor{33}$33.5$ & \heatcellcolor{22}$22.7$ & \heatcellcolor{8}$8.4$ \\
   \bottomrule
    \end{tabular}
    }
    \caption{\legendsquare{tableblue!40}~Reasoning model \legendsquare{tableorange!40}~Local model \\ Error distribution in the \pro benchmark. (A) is assonant, (B) is synonym, (C) is inverse and (D) is trivial. Values represent percentage scores.}
    \label{tab:errors_heatpmap}
\end{table}

In Table~\ref{tab:accuracy} we present the results recorded on the \pro benchmark and the ancillary tasks. Models are sorted based on their performance on the \pro task: such an ordering highlights a clear separation of performance between thinking \textit{vs.} non-thinking models. 

By comparing the performances between the ancillary tasks and the \pro benchmark, we uncover an unexpected phenomenon: virtually all non-thinking models suffer from steep performance deterioration. For instance, GPT 4o achieves 92\% on the \textit{base + true ending} task but only 64\% on \pro. Claude Sonnet 4 loses 47 percentage points, DeepSeek V3 loses 52 percentage points, and Grok 3 drops by 69 percentage points. The most dramatic performance decline occurs with LLama 4 Maverick, which plummets from 75\% and 88\% on the ancillary tasks to merely 6\% on \pro. Notably, Mistral's performance, given its relatively modest size (24B parameters), suggests that domain-specific optimization---through more focused Italian and broader European-language training~\cite{mistralhuggingface}---may play a significant role in enhancing model efficiency for culturally grounded tasks.

LRMs are less prone to this performance drop; however, we still observe significant deterioration of about 10-20 percentage points. These findings suggest that the transition from pattern completion to discriminative reasoning fundamentally challenges current language models' understanding mechanisms. The substantial performance gaps confirm that models rely heavily on memorized linguistic patterns rather than genuine semantic comprehension of proverbs. This deterioration becomes particularly pronounced when models must evaluate and reject plausible but incorrect alternatives, highlighting limitations in their ability to engage in deeper cultural and contextual reasoning. The relatively better performance of reasoning models suggests that explicit reasoning processes can partially compensate for these limitations, though significant challenges remain in achieving robust figurative language understanding.


\paragraph{Detailed error analysis.}
Table~\ref{tab:errors_heatpmap} details the categorization of incorrect responses as a percentage of total errors. 
The results reveal a strong skew toward option \textbf{B}, highlighting a consistent preference among the models for selecting synonyms--even if they are literal and not figurative. 
This pattern is less evident among local models, whose responses appear more equally distributed, possibly reflecting greater variability or reduced confidence in their outputs. The complete report of each model's responses is provided in Table~\ref{tab:errors_absolute} in the Supplementary Materials.


\subsubsection{CoTs Analysis}
\begin{figure*}[t]
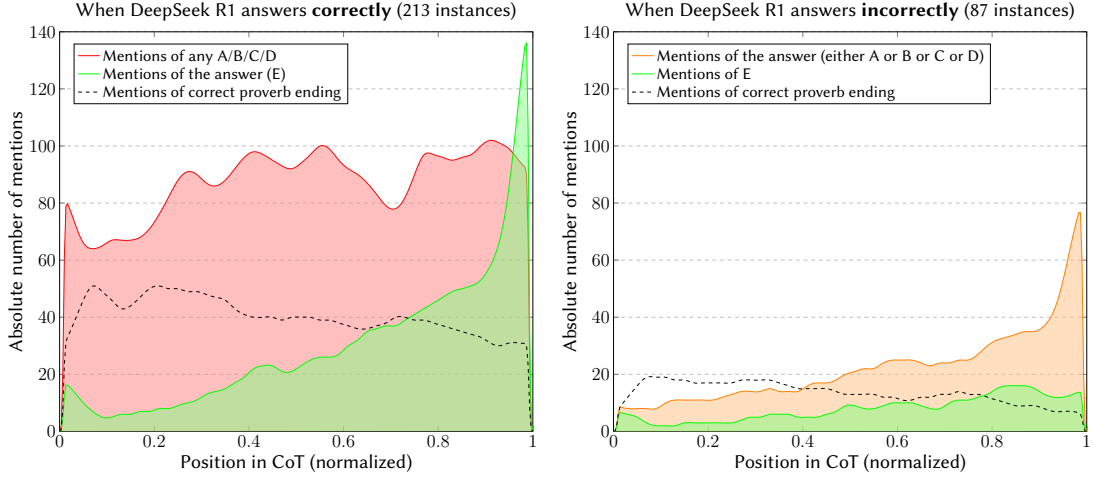

\centering
\resizebox{1\textwidth}{!}{
\begin{tabular}{cc}
\input{graphs/correct_response_analysis_r1} & \input{graphs/incorrect_response_analysis_r1} 
\end{tabular}
}
\caption{Analyses of the DeepSeek R1 CoTs. \textbf{Left} (the model answers correctly): tracing mentions of the \textcolor{green!70!black}{correct answer} (E) and \textcolor{red}{any incorrect} option (A/B/C/D). \textbf{Right} (the model answers incorrectly): tracing mentions of the \textcolor{green!70!black}{correct} (E) option and the exact provided \textcolor{orange}{incorrect answer} (either A or B or C or D). The dotted line shows the mentions of the true ending of the proverb (which was not given as option).}
\label{fig:cots-deepseek}
\end{figure*}

\begin{figure*}[t]
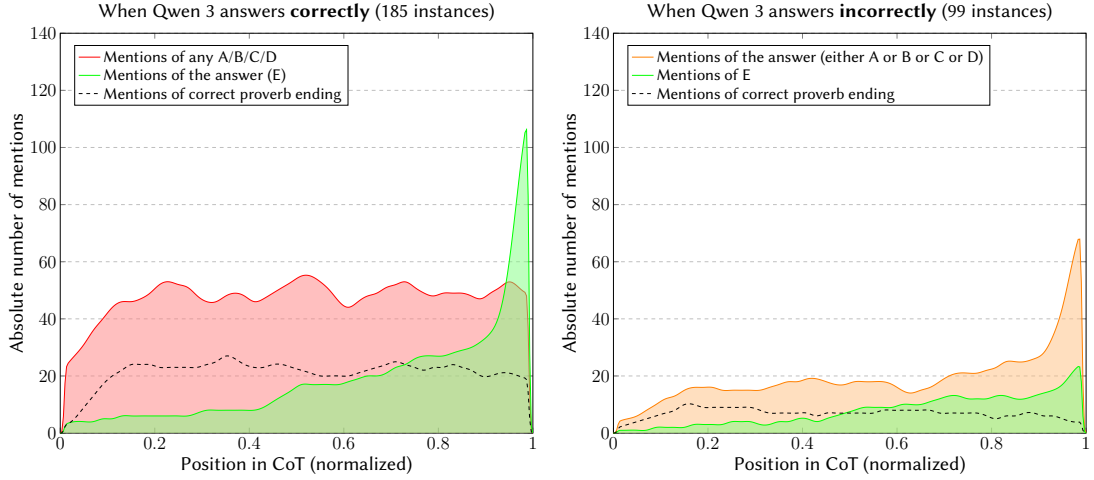

\centering
\resizebox{1\textwidth}{!}{
\begin{tabular}{cc}
\input{graphs/correct_response_analysis_qwen} & \input{graphs/incorrect_response_analysis_qwen}
\end{tabular}
}
\caption{Analyses of the Qwen 3 CoTs. \textbf{Left} (the model answers correctly): tracing mentions of the \textcolor{green!70!black}{correct answer} (E) and \textcolor{red}{any incorrect} option (A/B/C/D). \textbf{Right} (the model answers incorrectly): tracing mentions of the \textcolor{green!70!black}{correct} (E) option and the exact provided \textcolor{orange}{incorrect answer} (either A or B or C or D). The dotted line shows the mentions of the true ending of the proverb (which was not given as option).}\label{fig:cots-qwen}
\end{figure*}

For the CoT analysis, we only take in consideration DeepSeek~R1 and Qwen~3, as they are the only models that provide a full CoT trace rather than a summarization. As discussed earlier, these models were run with an unlimited thinking budget. 

Since we ran the \pro benchmark three times in order to compute the majority vote for the accuracy, we automatically analyzed a total of $600$ CoTs ($300$ for each model). Table~\ref{tab:cot_length} provides a preliminary overview of our analyses.
\begin{table}[h!]
\centering
\begin{tabular}{lcc}
\toprule
& DeepSeek R1 & Qwen 3 \\
\midrule
Analyzed CoTs & $300$ & $300$ \\
Empty CoTs & $0$ & $16$ ($5.33$\%) \\
Average Words & $796$ & $680$ \\
CoT > $2000$ Words & $7$ & $15$ \\
\midrule
Languages & \makecell{IT ($56$\%) \\ EN ($44$\%)} & \makecell{IT ($0$\%) \\ EN ($100$\%)} \\ 
\bottomrule
\end{tabular}
\caption{Overview of the Chain-of-Thought (CoT) analysis for DeepSeek R1 and Qwen 3.}
\label{tab:cot_length}
\end{table}
Most prompts provided a non-empty CoT, and from our investigation we discovered two interesting facts:
\begin{itemize}
    \item \textbf{Overthinking}: Models occasionally exhibit \textit{overthinking} behavior~\cite{sui2025stop}, a documented phenomenon affecting LRMs where they continuously re-evaluate their assessment of the correct answer. This results in CoTs exceeding 4,000 words in length, compared to an average of approximately 700 words for typical responses.
    \item \textbf{Language inconsistency in CoTs}:\footnote{Automatic language detection was performed via \url{https://pypi.org/project/langdetect/}.} Approximately half of DeepSeek's CoTs are generated in English while the other half appear in Italian, with occasional language switching occurring within individual reasoning traces. In contrast, Qwen consistently produces CoTs exclusively in English (except when citing the question options). This multilingual reasoning presents significant interpretability challenges, particularly for tasks involving idiomatic content, as cultural nuances and figurative meanings may be lost or misrepresented when reasoning shifts between languages~\cite{etxaniz2024multilingual, ranaldi2024limits}. We hypothesize that this limitation stems from these models' training distribution, which prioritizes Chinese and English content with comparatively limited Italian language exposure.
\end{itemize}

We analyzed the non-empty CoTs by tracing mentions of correct and incorrect answers within the thinking process. We examined separately cases where the model responds correctly versus incorrectly. Specifically, the left subfigures of Figures~\ref{fig:cots-deepseek} and~\ref{fig:cots-qwen} show the absolute number of mentions of the correct answer (which is always E - \textit{None of the others}) and all incorrect answers when the model answers correctly. Conversely, the right subfigures show the absolute number of mentions of the correct answer and the specific incorrect answer provided when the model responds incorrectly. We additionally plot as a dotted line the absolute number of mentions of the correct completion of the proverb (which was not given in the prompt).

These graphs reveal that both models continuously mention all possible answers throughout their reasoning process, while the spikes toward the end indicate that models reach a decision only in the final lines of their CoTs. However, this decision-making appears tentative, as alternative options remain heavily mentioned alongside the chosen answer, suggesting low confidence in the final selection.

The dotted lines clearly demonstrate that models are aware of the correct proverb ending and repeatedly reference it throughout their thinking process. These observations highlight a critical disconnect: while models can successfully recall the correct proverb completion, they fail to recognize its absence among the provided choices. This suggests that the challenge lies not in knowledge retrieval but in the discriminative reasoning required to identify when the correct answer is unavailable, revealing fundamental limitations in how current LRMs handle negative reasoning tasks~\cite{salido2025none}.

\paragraph{Inconsistency between CoTs and answers.} As a final finding, we also discovered that sometimes model responses were inconsistent with their corresponding CoT. For instance, out of the $113$ incorrect responses from Qwen, $14$ of them are inconsistent, ending with sentences like \textit{`The answer is \textbf{X}'}, but then the actual given answer was not \textbf{X}. Remarkably, in all of these instances adhering to the CoT-delivered conclusion would have resulted in a correct answer. Similarly, for DeepSeek~R1, $6$ of the $87$ incorrect responses exhibited such a discrepancy, $5$ ($5.7$\%) of which would have lead to the correct answer. This behavior has been observed in prior work~\cite{chen2025reasoning}.

In the Supplementary Materials we report two complete CoTs showing instances of english that leads to a wrong answer and answer mismatch.

\section{Conclusions \& Future Work}
\label{sec:conclusions}

In this work, we introduced \pro, a novel Italian benchmark designed to evaluate Large Language Models' ability to handle culturally grounded linguistic expressions through proverb completion tasks. Our comprehensive evaluation of $13$ frontier models, including both Large Reasoning Models and traditional LLMs, provides significant insights into the limitations of current language understanding systems.

Our findings demonstrate a relevant gap between models' knowledge of proverbs and their ability to apply this knowledge in discriminative reasoning tasks. While nearly all evaluated models successfully complete proverbs when presented with direct completion prompts, performance drops dramatically when the same task is reformulated as multiple-choice selection without correct answers available. Even state-of-the-art reasoning models like GPT o3 and Gemini 2.5 Pro experience substantial degradation. 

The Chain-of-Thought analysis of DeepSeek R1 and Qwen 3 further highlights this limitation: both models frequently mention correct proverb endings during their reasoning process yet fail to recognize their absence from the provided options, highlighting fundamental challenges in negative reasoning capabilities. Moreover, we uncovered concerning inconsistencies in reasoning model behavior, including overthinking, language switching during reasoning and discrepancies between CoT conclusions and final answers.

Future work should focus on investigating this mismatch between knowledge retrieval and discriminative reasoning more deeply, particularly examining how models handle negative reasoning tasks even in seemingly trivial scenarios where the correct answer is absent from the given options. Additional evaluation methodologies should also be incorporated, including answer randomization techniques as proposed in literature~\cite{wang2024my}.

In summary, our results underscore the critical importance of developing language-specific benchmarks that capture cultural and linguistic nuances often lost in English-centric evaluations, showing that current LLMs rely heavily on memorized patterns rather than deeper semantic understanding of culturally grounded expressions, highlighting important limitations in their reasoning capabilities for figurative language comprehension.

%
%
%
%
%
%
%
%
%


\bibliography{ceur}




\appendix
\clearpage
\section*{Supplementary Materials}
\label{sec:supp}
\vspace{1em} 

\subsection*{Proverb Examples}
In the following, we report a few examples in Italian from the \pro benchmark. Each example presents the beginning of a proverb followed by five possible continuations. Only one option is correct, which is \textit{E) Nessuna delle altre risposte} (None of the other answers).

\begin{styledtext}{Example 1}%
Non è tutto oro... \\
A) ...quel che lucida \\
B) ...ciò che brilla \\
C) ...ciò che è opaco \\
D) ...quel che è metallico \\
E) Nessuna delle altre risposte
\end{styledtext}

\begin{styledtext}{Example 2}%
Vivi e ... \\
A) ...lascia ridere \\
B) ...fai decorrere \\
C) ...giudica tutti \\
D) ...morirai più tardi \\
E) Nessuna delle altre risposte
\end{styledtext}

\begin{styledtext}{Example 3}%
Il gioco è bello quando... \\
A) ...cura dopo \\
B) ...è breve \\
C) ...dura molto \\
D) ...tutti si divertono \\
E) Nessuna delle altre risposte
\end{styledtext}

\begin{styledtext}{Example 4}%
A caval donato ... \\
A) ...non si guarda la brocca \\
B) ...non si ispezionan le fauci \\
C) ...si guarda in bocca \\
D) ...si mette la sella \\
E) Nessuna delle altre risposte
\end{styledtext}

\begin{styledtext}{Example 5}%
Chi non beve in compagnia... \\
A) ...o è un lardo o è una spa \\
B) ...o è un delinquente o è uno spione \\
C) ...non è un ladro né una spia \\
D) ...guida la macchina \\
E) Nessuna delle altre risposte
\end{styledtext}

\begin{styledtext}{Example 6}%
Chi dorme...\\
A) ...non piglia cesti\\
B) ...non acchiappa trote\\
C) ...piglia pesci\\
D) ...non è sveglio\\
E) Nessuna delle altre risposte
\end{styledtext}

\newpage 
\subsection*{Complete CoTs}
\label{sec:complete_cots}
In the following, we report two examples of CoT. In the first, the use of English leads to an incorrect answer, while in the second, an answer mismatch is observed: although the CoT concludes with option C, the model selects option B.

\begin{styledtexttt}{CoT of Qwen\\The use of English leads to a wrong answer}%
Okay, let's see. The user wants me to complete the proverb "Lontano dagli occhi,..." by choosing the correct option from A to E. The options are: A) ...lontano dalle suore, B) ...distante dal miocardio, C) ...vicino al cuore, D) ...si vede poco, E) Nessuna delle altre risposte.

First, I need to recall the actual Italian proverb. The common one I know is "Lontano dagli occhi, vicino al cuore," which translates to "Out of sight, out of mind" but literally means "Far from the eyes, close to the heart." Wait, that doesn't make sense in terms of translation, but the structure is similar to English proverbs where the second part contrasts or complements the first. So the original proverb should have a contrasting part. In Italian, "vicino al cuore" is the correct completion, meaning that even if someone is far from your eyes (not seen), they are still near your heart (in your thoughts or loved).

Now, checking the options. Option C is "...vicino al cuore," which matches what I know. The other options: A) "lontano dalle suore" (far from the nuns) - doesn't make sense in this context. B) "distante dal miocardio" (distant from the myocardium) - that's a medical term, not a proverb. D) "si vede poco" (you see little) - logical but not the traditional saying. E) would only be correct if none of the options are right, but since C seems correct, E isn't the choice here. So the answer must be C.
\tcblower
C. 
\end{styledtexttt}

\begin{styledtexttt}{CoT of DeepSeek \\ Answer Mismatch w.r.t the CoT }%
Okay, ho bisogno di completare correttamente il proverbio "Non è tutto oro..." scegliendo tra le opzioni date. Le opzioni sono: A) ...quel che lucida B) ...ciò che brilla C) ...ciò che è opaco D) ...quel che è metallico E) Nessuna delle altre risposte

Prima, mi ricordo come va il proverbio completo. So che il proverbio classico è "Non è tutto oro quel che luccica". Tuttavia, nessuna delle opzioni offre esattamente "luccica".

Guardando le opzioni: A) "quel che lucida" - "Lucida" è simile a "luccica", dato che entrambi significano brillare o splendere. B) "ciò che brilla" - Anche questo significa brillare, quindi è molto vicino al significato originale. C) "ciò che è opaco" - Opaco è l'opposto di brillante, quindi non ha senso in questo contesto. D) "quel che è metallico" - Mentre l'oro è un metallo, il proverbio non si riferisce semplicemente ai metalli ma a ciò che appare prezioso ma non lo è.

L'essenza del proverbio è che non tutto ciò che sembra prezioso o attraente (come l'oro che brilla) è effettivamente di valore. Quindi le opzioni A e B catturano entrambe questo significato.

Confrontando A e B: "quel che lucida" usa "lucida", che è un sinonimo più diretto di "luccica" nel proverbio originale. "Lucida" significa lucidare o brillare, quindi ha lo stesso senso. Invece, "ciò che brilla" (B) è più generico per brillare, ma comunque corretto nel contesto. 

Tuttavia, il proverbio originale italiano è tipicamente "Non è tutto oro quel che luccica", quindi "quel che lucida" (A) è la versione più fedele perché "lucida" funziona come sostituto di "luccica".

L'opzione E ("Nessuna delle altre risposte") non è necessaria perché A o B sono accettabili. Ma devo vedere quale è la scelta migliore.

Analizzo: Tecnicamente, la versione corretta è con "luccica", ma poiché non è tra le opzioni, "lucida" è un'alternativa valida e comune. Inoltre, molte fonti a volte usano varianti come "brilla", ma "lucida" è più vicino in termini di espressione.

Ho sentito usare sia "luccica" che "lucida" informalmente, quindi A mi sembra corretta. B è anche corretta nel significato, ma non è la formulazione tradizionale. Quindi scelgo A come risposta.

\tcblower
B. 
\end{styledtexttt}

\newpage

\begin{table*}[h]
  \centering
  \label{tab:errors_absolute}
  \begin{tabular}{lccccc}
    \toprule
    Model & A) Assonant & B) Synonym & C) Inverse & D) Trivial & E) None of the others\\
    \midrule
    \rowcolor{tableblue!40} GPT o3 & 2 & 34 & 3 & 0 & 261 \\
    \rowcolor{tableblue!40} Gemini 2.5 Pro & 3 & 62 & 5 & 1 & 229 \\
    \rowcolor{tableblue!40} DeepSeek R1 & 2 & 80 & 2 & 3 & 213 \\
    \rowcolor{tableblue!40} Claude Sonnet 4 & 3 & 77 & 10 & 0 & 210 \\
    \rowcolor{tableblue!40} Qwen 3 & 12 & 63 & 32 & 6 & 187 \\
    GPT 4o & 3 & 78 & 18 & 5 & 196 \\
    Claude Sonnet 4 & 3 & 101 & 50 & 8 & 138 \\
    DeepSeek V3 & 25 & 89 & 51 & 20 & 115 \\
    Grok 3 & 16 & 162 & 36 & 4 & 82 \\
    Gemini 2.5 Flash & 19 & 136 & 82 & 25 & 38 \\
    LLama 4 Maverick & 50 & 149 & 60 & 26 & 15 \\
    \rowcolor{tableorange!40} Mistral Small 3.1& 19 & 86 & 56 & 46 & 93 \\
    \rowcolor{tableorange!40} Gemma 3 & 102 & 96 & 65 & 24 & 13 \\
    \bottomrule
  \end{tabular}
    \caption{\legendsquare{tableblue!40}~Reasoning model \legendsquare{tableorange!40}~Local models \\ Absolute number of responses for each error type in the \pro task.}
\end{table*}

\end{document}